\documentclass{article}

\usepackage{graphicx} 
\usepackage{subfigure}

\usepackage{natbib}

\usepackage{algorithm}
\usepackage{algorithmic}

\usepackage{ams math}
\usepackage{amsfonts}

\usepackage{hyperref}

\newcommand{\R}{\mathcal{R}} 
\newcommand{\F}{\mathcal{F}} 
\newcommand{\M}{\mathcal{M}_{\F,\R}} 
\newcommand{\G}{\mathcal{G}_{\F,\R}} 

\usepackage[accepted]{icml2013mod}

\icmltitlerunning{The Weighted Stochastic Block Model}

\begin{document} 

\twocolumn[
\icmltitle{Adapting the Stochastic Block Model to Edge-Weighted Networks}

\icmlauthor{Christopher Aicher}{christopher.aicher@colorado.edu}
\icmladdress{Department of Applied Mathematics, University of Colorado, Boulder, CO 80309}
\icmlauthor{Abigail Z.\ Jacobs}{abigail.jacobs@colorado.edu}
\icmladdress{Department of Computer Science, University of Colorado, Boulder, CO 80309}
\icmlauthor{Aaron Clauset}{aaron.clauset@colorado.edu}
\icmladdress{Department of Computer Science and BioFrontiers Institute, University of Colorado, Boulder, CO 80309\\ 
Santa Fe Institute, Santa Fe, NM 87501}

\icmlkeywords{block model, exponential family, weighted relational data, variational Bayes}

\vskip 0.3in
]

\begin{abstract} 
We generalize the stochastic block model to the important case in which edges are annotated with weights drawn from an exponential family distribution. This generalization introduces several technical difficulties for model estimation, which we solve using a Bayesian approach. We introduce a variational algorithm that efficiently approximates the model's posterior distribution for dense graphs. In specific numerical experiments on edge-weighted networks, this \textit{weighted stochastic block model} outperforms the common approach of first applying a single threshold to all weights and then applying the classic stochastic block model, which can obscure latent block structure in networks. 
This model will enable the recovery of latent structure in a broader range of network data than was previously possible.
\end{abstract} 

\section{Introduction}

In social and biological networks, vertices often play distinct functional roles in the large-scale structure of the graph. The automatic detection of these latent roles, by identifying the induced ``community'' or block structures from connectivity data alone, is a fundamental problem in network analysis and many approaches have been proposed~\cite{fortunato:2010, porter_communities_2009}. The stochastic block model (SBM) is a popular generative model that solves this problem in an unsupervised fashion~\cite{holland:laskey:leinhardt:1983, wang:wong:1987}.




In its classic form, the SBM is a probabilistic model of pairwise interactions among $n$ vertices. Each vertex belongs to one of $k$ latent groups, and each undirected edge exists or does not with a probability that depends only on the block memberships of the connecting vertices. The model is thus defined by a vector $z$ containing the block assignment of each vertex and a $k\times k$ matrix $p$, where $p_{uv}$ gives the probability that a vertex of block $u$ connects to a vertex of block $v$.

This model can capture a wide variety of large-scale organizational patterns of network connectivity, depending on the choices of $p$ and $z$. If $p$'s diagonal elements are greater than its off-diagonal elements, the block structure is assortative, with communities exhibiting greater edge densities within than between them, as is often found in social networks. Other choices of $p$ can generate hierarchical, multi-partite, or core-periphery patterns, among others. This flexibility, and the principled probabilistic statements it produces, has made the SBM a popular tool for unsupervised network analysis, in which we seek to infer the latent block labels from the observed graph structure alone.

There is broad interest in machine learning, physics, and computational social science to develop and apply generalizations of the classic SBM. Generalizations have been made to allow degree heterogeneity within blocks~\cite{karrer_stochastic_2011}, probabilistic or mixed block membership~\cite{airoldi_mixed_2008,ball:karrer:newman:2011}, infinite number of blocks~\cite{kemp_irm_2006}, or hierarchical (nested) relationships among blocks~\cite{clauset:moore:newman:2008}. 

 
Several efficient techniques exist for estimating latent block structures from data. Of particular relevance to our weighted generalization of the SBM are the variational algorithms, both Bayesian and frequentist. Scalability is typically achieved by constraining the parameter space or using modern optimization techniques. Examples include variational expectation-maximization (EM) for the classic SBM~\cite{daudin_mixture_2008, park_dynamic_2010}, variational Bayes EM for a restricted, two-parameter $p$ matrix~\cite{hofman_bayesian_2008}, nested variational EM for the classic mixed membership SBM~\cite{airoldi_mixed_2008}, and stochastic variational inference for assortative mixed membership SBM~\cite{gopalan_scalable_2012}. 


In most of these efforts, the SBM is restricted to binary or Bernoulli networks, in which edges are unweighted. The one exception has been block models with Poisson distributed edge weights~\cite{mariadassou_variationalweight_2010,karrer_stochastic_2011,ball:karrer:newman:2011}, which can be fitted to multigraphs.
In practice, however, most binary networks are produced after applying a threshold to a weighted relationship~\cite{thomas_valued_2011}, and this practice clearly destroys potentially valuable information. To apply the SBM on weighted data without thresholding, we introduce a generalization of the SBM to the important case in which edges are annotated with weights drawn from an exponential family distribution. 

This \textit{weighted stochastic block model} (WSBM) includes as special cases most standard distributional forms, and thus allows us to use weighted relations directly in recovering latent block structure, preventing the information loss caused by thresholding. Handling these general weight distributions presents several technical difficulties for model estimation, which we solve using a Bayesian approach. We first give the WSBM's form and derive a variational Bayes algorithm for fitting to dense graphs. We then present synthetic examples that illustrate the type of behavior the WSBM captures that is overlooked by thresholding. We close with a brief discussion of extensions of the model. 


\section{Weighted Stochastic Block Models}

The weighted stochastic block model is a generative model for weighted pairwise interactions among $n$ vertices, and is composed of an exponential family distribution $\F$ and a block structure $\R$. The block structure defines a set of vertex labels, denoted $z = \{z_1,\ldots,z_n\}$ where $z_i \in K = \{1,\ldots,k\}$. The block structure $\mathcal{R}$ defines a partition on the edges into $R$ disjoint \textit{bundles}, one for each pair of blocks. Edges weights in some bundle are modeled by a distribution in $\F$, 
parameterized by $\theta_r \in \theta = \{\theta_1, \ldots, \theta_R\}$. That is, each bundle has its own set of distribution parameters.

The choice of $\R$ determines the large-scale structure of the network, just as $p$ and $z$ do for the classic SBM.
When $\F$ is a Bernoulli trial, we cover this classic case. Although constraining $\R$, or the variation of its parameters across edge bundles, can be used to create specific types of large-scale structure, here we focus on the general case of blocks with independent parameters.
In principle, the form of $\R$ could be 
learned directly from data, but we do not explore this topic. 

We denote a WSBM with edge distribution family $\F$ and block structure $\R$ by $\M$, whose parameters are the vertex labels $z$ and the matrix of edge bundle parameters $\theta$. The likelihood of observing a graph $A$, given distribution $f \in \F$, is then 
\begin{equation*}
\Pr(A \, | \, z,\theta, \M) = \prod_{i<j} f(A_{i,j} \, | \, \theta_{\R(z_i,z_j)}) \enspace .
\end{equation*}
%

Restricting $\F$ to exponential family distribution makes the mathematics tractable while covering a broad range of models of edge weights, including many common distributions produced by classic stochastic processes.
A distribution $f$ belongs to an exponential family $\F$ if it can be written as
\begin{equation*}
f(x\,|\,\phi) = h(x) \exp\left( T(x) \cdot \eta(\phi) \right) \text{ for } x \in \mathcal{X} \enspace
\end{equation*}
where $h$, $T$, $\eta$ are fixed mappings, $\phi$ is the distribution's parameter, and $\mathcal{X}$ is the distribution's support. Under these assumptions, the log-likelihood becomes
\begin{equation*}
\mathcal{L} = \sum_{i < j} \log h(A_{i,j}) + \sum_{r = 1}^R T_r \cdot \eta(\theta_r) \enspace
\end{equation*}
where $T_r = \sum_{i,j : \R(z_i,z_j) = r} T(A_{i,j})$ is the sufficient statistic for the weights in edge bundle $r$.

For some choices of $\M$, the likelihood function contains degeneracies that prevent the direct estimation of parameters $z$ and $\theta$. For instance, when weights are real-valued and $\F$ is a Normal distribution. An edge bundle with all-equal weights will have zero variance, which creates a degeneracy in the likelihood calculation. 
Another technical problem is that non-edges in a sparse graph (a zero in the adjacency matrix) may represent a pair of non-interacting vertices, an interaction with zero weight, or an interaction we have not yet observed.
The classic SBM does not exhibit these problems because edge weights are Bernoulli random variables, whose sufficient statistics are always well defined.
To regularize the degeneracy problem, we take a Bayesian approach and assign an appropriate prior distribution $\pi$ to our parameters $\theta$. Now, the posterior distribution $\pi^{*}$ will exhibit no degeneracies and estimation can proceed smoothly. 

Estimating the posterior distribution $\pi^{*}(z,\theta\,|\,A)$ given the observed edge weights $A$ and prior $\pi$ is generally difficult, and so we approximate $\pi^{*}$ by a factorizable distribution $q(z,\theta) = q(z)q(\theta)$.
How we estimate $\pi^{*}$ also depends on whether the graph $A$ is dense or sparse, 
and our interpretation of non-edges. Here, we present the solution for dense graphs. In a separate paper, we will present a belief propagation algorithm for sparse graphs that correctly handles non-edges.

\section{Variational Bayes} 

For a dense graph, we construct a variational Bayes (VB) expectation-maximization algorithm to estimate $\pi^{*}$.
We approximate the posterior distribution $\pi^{*}(z,\theta|A)$ by a product of marginals $q(z,\theta) = \prod_i q_i(z_i)\prod_r q(\theta_r)$.

We then select $q$ by minimizing the Kullback-Leibler (KL) divergence between our approximation and the posterior $D_{\text{KL}}(q\, || \,\pi^{*})$.
It can be shown that
\begin{equation*}
\log \Pr(A\, |\,\M) = \G(q) + D_{\text{KL}}\left( q \,||\, \pi^{*}\right) \enspace ,
\end{equation*}
where $\G(q)$ is a functional lower bound on the constant $\log \Pr(A \ | \ \M)$, calculated as
\begin{equation*}
\G(q) = \mathbb{E}_{q}\left(\mathcal{L}\right) + \mathbb{E}_{q}\!\left(\log \frac{\pi( z , \theta)}{q( z , \theta)}\right) \enspace .
\end{equation*}
The first term is the expected log-likelihood under the approximation $q$ and the second term is the KL-divergence of the approximation $q$ from the prior $\pi$. As the likelihood $\log \Pr(A\, |\, \M)$ is constant, minimizing the KL divergence $D_{\text{KL}}(q\, || \,\pi^{*})$ is equivalent to maximizing $\G(q)$. 

To maximize $\G$, we maximize the expected log-likelihood of the data and weakly constrain the approximation to be close to the prior. This 
regularizer prevents over fitting and eliminates the aforementioned likelihood degeneracies. In practice, the first term overwhelms the second term given sufficient data.

\paragraph{Conjugate priors.}
For mathematical convenience, we restrict the prior $\pi$ to a product of parameterized conjugate distributions.

The conjugate prior for the parameter $\theta$ of an exponential family has the form
\begin{equation*}
\pi(\theta) = Z^{-1}(\tau)\exp\left(\tau \cdot \eta(\theta)\right) \enspace ,
\end{equation*}
where $\tau$ parameterizes the prior and $Z(\tau)$ is a normalizing constant. When we update the prior based on the observed weights in a given edge bundle $r$, the posterior's parameter becomes $\tau^* = \tau + T_r$, 
and $\tau$ can be viewed as a set of pseudo-observations. This prevents the posterior from becoming degenerate since every edge bundle, no matter how small or uniform, produces a parameter estimate.

The conjugate prior for a vertex label $z$ is a categorical distribution with parameter $\mu \in \mathbb{R}^k$, where $\mu_i(\kappa)$ is the probability that node $i$ belongs to group $\kappa$. We fit $\mu_i$ directly, with a flat prior $\mu_0(\kappa) = 1/k$.


The form of our prior is thus
\begin{equation*}
\pi(z, \theta) = \prod_i \mu_0(z_i) \prod_{r}  Z^{-1}(\tau_0)\exp\left(\tau_0 \cdot \eta(\theta_r)\right)  \enspace ,
\end{equation*}
where $\mu_0$, $\tau_0$ are the parameters for the priors $\pi_i$, $\pi_r$. With conjugate priors for $\pi$, our approximation $q$ takes the form
\begin{equation*}
q(z, \theta) = \prod_i \mu_{i}(z_i) \prod_{r}  Z^{-1}(\tau_r)\exp\left(\tau_r \cdot \eta(\theta_r)\right)\enspace .
\end{equation*}
Now, maximizing $\G$ is equivalent to maximizing $\G$ over $q$'s parameters $\mu_i$, $\tau_r$. 

\paragraph{Optimizing $\G$.} 
These choices of $\pi$ and $q$ yield
%
\begin{align*} \G &= \sum_{i,j} \log h(A_{i,j}) + \sum_r \left(\left\langle T \right\rangle_r + \tau_0 - \tau_r \right) \cdot \left\langle \eta \right\rangle_r  \\
 &\quad + \sum_r \log \frac{Z(\tau_r)}{Z(\tau_0)} + \sum_i \sum_{z_i} \mu_i (z_i) \log \frac{\mu_0 (z_i)}{\mu_i (z_i)} \enspace ,
 \end{align*} 
where $\left\langle T \right\rangle_r$, $\left\langle \eta \right\rangle_r$ are expectations of $T_r$, $\eta_r$ under the approximation $q$; for exponential families they are,
\begin{align*}
  \left\langle T \right\rangle_r & := \sum_{i,j} \sum_{R(z_i,z_j) = r} \mu_i(z_i) \, \mu_j(z_j) \, T(A_{i,j}) \nonumber \\
  \left\langle \eta \right\rangle_r & := \frac{\partial \log Z(\tau_r)}{\partial \tau_r} \enspace . \nonumber
\end{align*}

To optimize $\G$ we take derivatives with respect to $q$'s parameters $\mu$, $\tau$ and set them to zero. We iteratively solve for the maximum by updating $\mu$ and $\tau$ independently.

For $\tau$, this yields
\begin{align*}
\frac{\partial \G}{\partial \tau_r} &= \left[\left\langle T \right\rangle_r + \tau_0 - \tau_r\right] \frac{\partial \left\langle \eta \right\rangle_r}{\partial \tau_r} - \left\langle \eta \right\rangle_r + \frac{\partial \log Z_r}{\partial \tau_r} \\
 &\propto \left\langle T \right\rangle_r + \tau_0 - \tau_r \enspace ,
\end{align*}
and the update equation for each edge-bundle parameter is $\tau_r = \tau_0+ \left\langle T \right\rangle_r  $.

For $\mu$, we use Lagrange multipliers $\lambda_i$ to enforce $\sum_{z} \mu_i(z) = 1$. Setting the derivative of $\G$ with respect to $\mu_i$ equal to $\lambda_i$ yields
 \begin{equation*}
\frac{\partial \G}{\partial \mu_i(z)} = \sum_r \left[ \frac{\partial \left\langle T \right\rangle_r}{\partial \mu_i(z)} \cdot \left\langle \eta \right\rangle_r \right] - \log \mu_i(z) = \lambda_i \enspace ,
\end{equation*}
where
\begin{equation*}
\frac{\partial \left\langle T \right\rangle_r}{\partial \mu_i(z)} := \sum_{z' : R(z,z') = r} \sum_{j \neq i} T(A_{i,j}) \mu_j(z') \enspace .
\end{equation*}
Solving for $\mu_i(z)$ produces the update equation
\begin{equation*}
\mu_i(z) \propto \exp\! \left( \sum_r \frac{\partial \left\langle T \right\rangle_r}{\partial \mu_i(z)} \cdot \left\langle \eta \right\rangle_r \right) \enspace ,
\end{equation*}
where each $\mu_i$ is normalized to a probability distribution. To calculate the $\mu_i$ values, we iteratively update each $\mu_i$ from some initial guess until convergence to within some tolerance.

\begin{algorithm}[tb]
   \caption{VB for dense networks}
   \label{alg:VB}
\begin{algorithmic}
   \STATE {\bfseries Input:} Data $A$, Model $\M$
   \STATE Initialize $\mu$
   \REPEAT
   \FORALL{$r = 1,\ldots, R$}
   \STATE Set $\left\langle T \right\rangle_r := \sum_{i,j} \sum_{\mathcal{R}(z_i,z_j) = r} \mu_i(z_i) \mu_j(z_j) T(A_{i,j})$
   \STATE Set $\tau_r := \tau_0 + \left\langle T \right\rangle_r$
   \STATE Set $\left\langle \eta \right\rangle_r := \left. \frac{\partial}{\partial \tau} \log Z(\tau) \right|_{\tau=\tau_r}$ 
   \ENDFOR
   \REPEAT
   \FORALL{$i = 1,\ldots, n$}
   \STATE $\frac{\partial \left\langle T \right\rangle_r}{\partial \mu_i(z)} := \sum_{\mathcal{R}(z,z') = r} \sum_{j \neq i} T(A_{i,j}) \mu_j(z') $
   \STATE $ \mu_i(z) \propto \exp \left( \sum_r \frac{\partial \left\langle T \right\rangle_r}{\partial \mu_i(z)} \cdot \left\langle \eta \right\rangle_r \right) $
   \ENDFOR 
   \UNTIL{$\mu$ converge}
   \UNTIL{$\mu,\tau$ converge }
   \RETURN $\mu,\tau$
\end{algorithmic}
\end{algorithm}

Algorithm~\ref{alg:VB} gives pseudocode for the full variational Bayes algorithm, which alternates between updating the edge-bundle parameters and the vertex label parameters using the update equations derived above. Because every pairwise interaction contributes to the estimation of some parameter, the algorithm takes $O(n^2)$ time, assuming fast convergence on $\theta$ and $\mu$.
%
%
Like all VB approaches, only convergence to a local optima of $\G$ is guaranteed. In practical contexts, multiple trials with a variety of initial conditions are used, and the best overall model selected.
\section{Model Selection} 

An important intermediate step toward applying the WSBM to some graph is the selection of a class of distributions $\F$ or the number of blocks $k$. Any of a number of principled approaches could be employed, including maximum likelihood, possibly with cross-validation~\cite{airoldi_mixed_2008}, Bayes factors~\cite{hofman_bayesian_2008}, approximations thereof~\cite{mariadassou_variationalweight_2010, daudin_mixture_2008}, or minimum description length~\cite{peixoto_2012_parsimonious}.

In our experiments below, we use Bayes factors, which assume a uniform prior and are equivalent to selecting the model with the largest model-likelihood,
\begin{equation*}
\log B(\mathcal{M}_1,\mathcal{M}_2) = \log\frac{\Pr(A \,|\, \mathcal{M}_1)}{\Pr(A \,|\, \mathcal{M}_2)} \approx {\G}_1 - {\G}_2 \enspace ,
\end{equation*}
where we approximate $\log\Pr(A \,|\, \M)$ with $\G$.

Although Bayes factors assign a uniform prior on a set of nested models, they have a built-in penalty for complex models. Recall that $\G$ is penalized for large divergence from the prior and since the vertex-label prior is uniform on all $k$ groups, there is a penalty if an increase in $k$ does not sufficiently reduce the entropy or correspondingly increase the expected log-likelihood.




\section{Experimental results}
%

\begin{figure*}[ht]         
	\centering
		\subfigure[Example Plot]{
			\raisebox{.18\height}{\includegraphics[width = .23\textwidth]{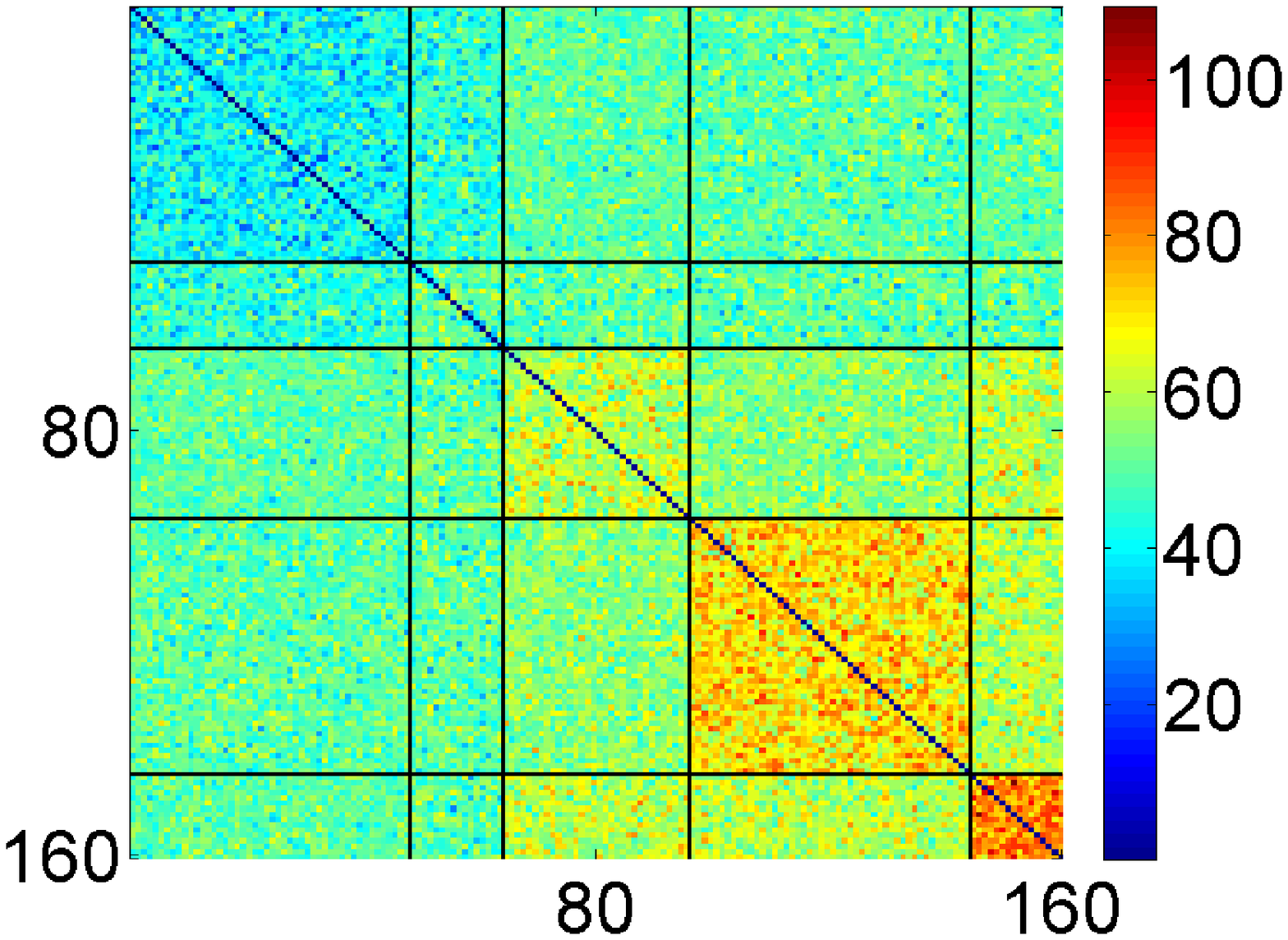}}
			\label{fig:1}
			}
		\subfigure[{Normal VI vs.\ $k$}]{
			\includegraphics[width = .23\textwidth]{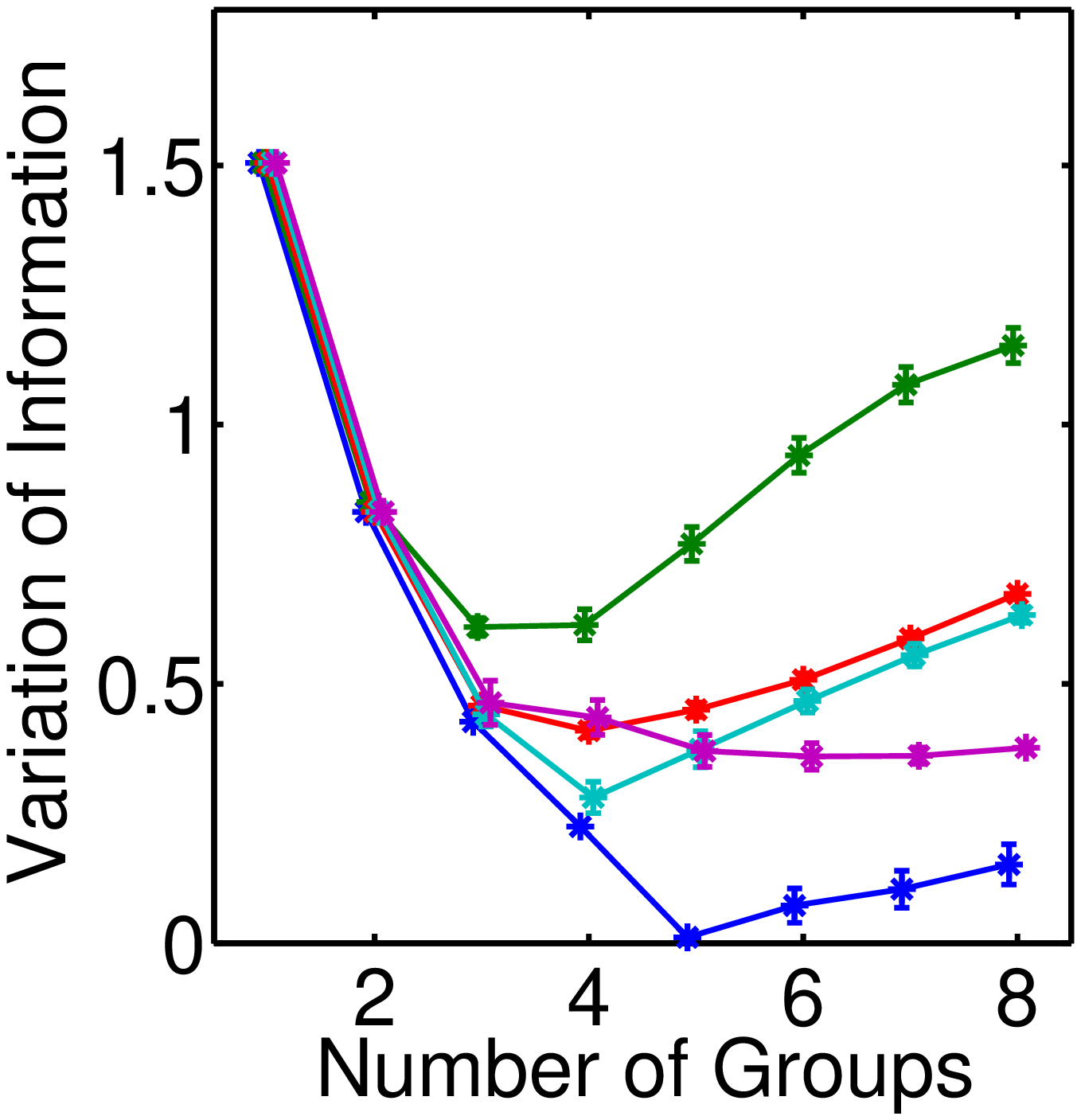}
			\label{fig:2}
			}
		\subfigure[{Normal VI vs.\ Variance}]{
			\includegraphics[width = .23\textwidth]{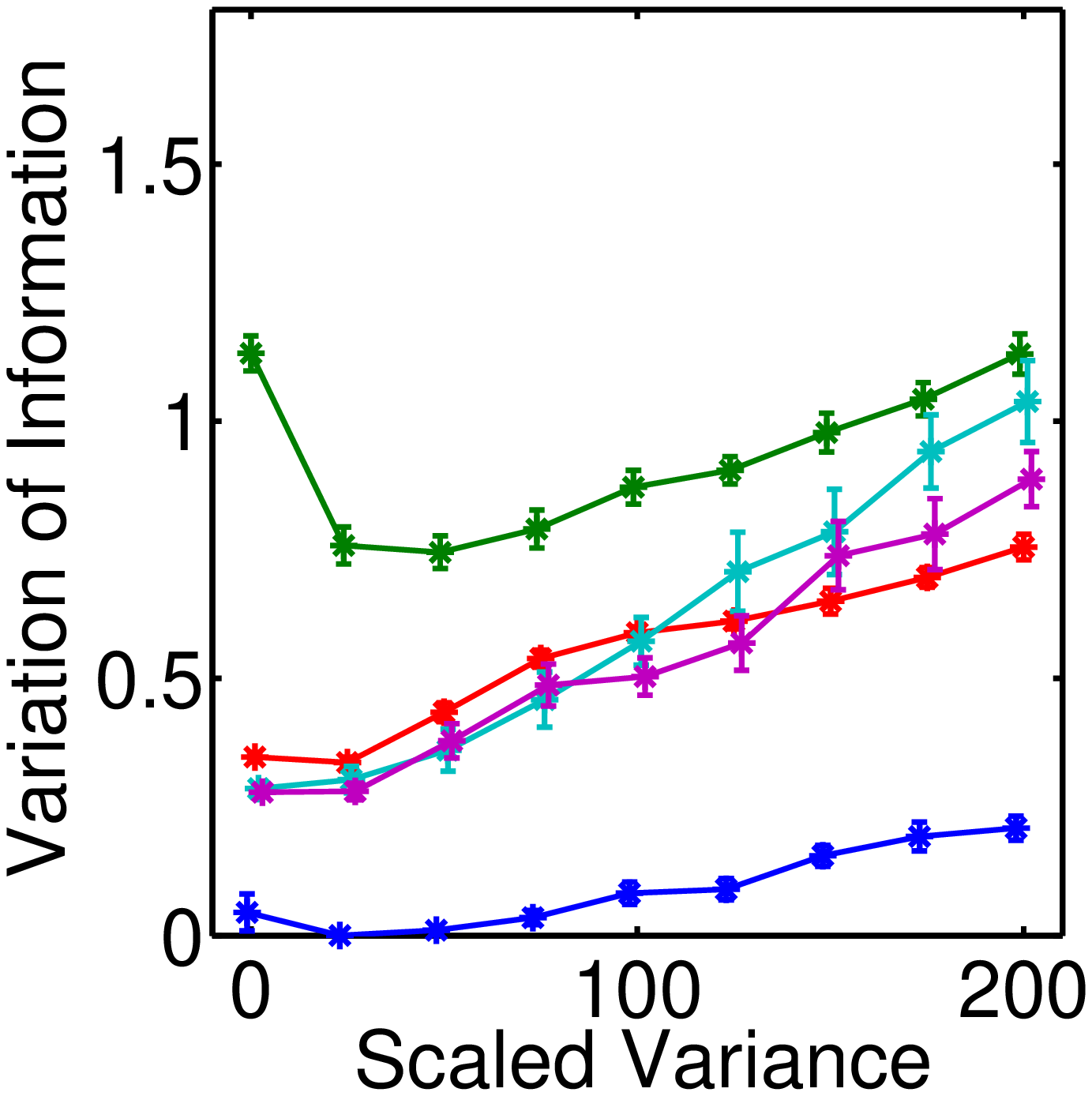}
			\label{fig:3}
			}
		\subfigure[{Normal VI vs.\ $n$}]{
			\includegraphics[width = .23\textwidth]{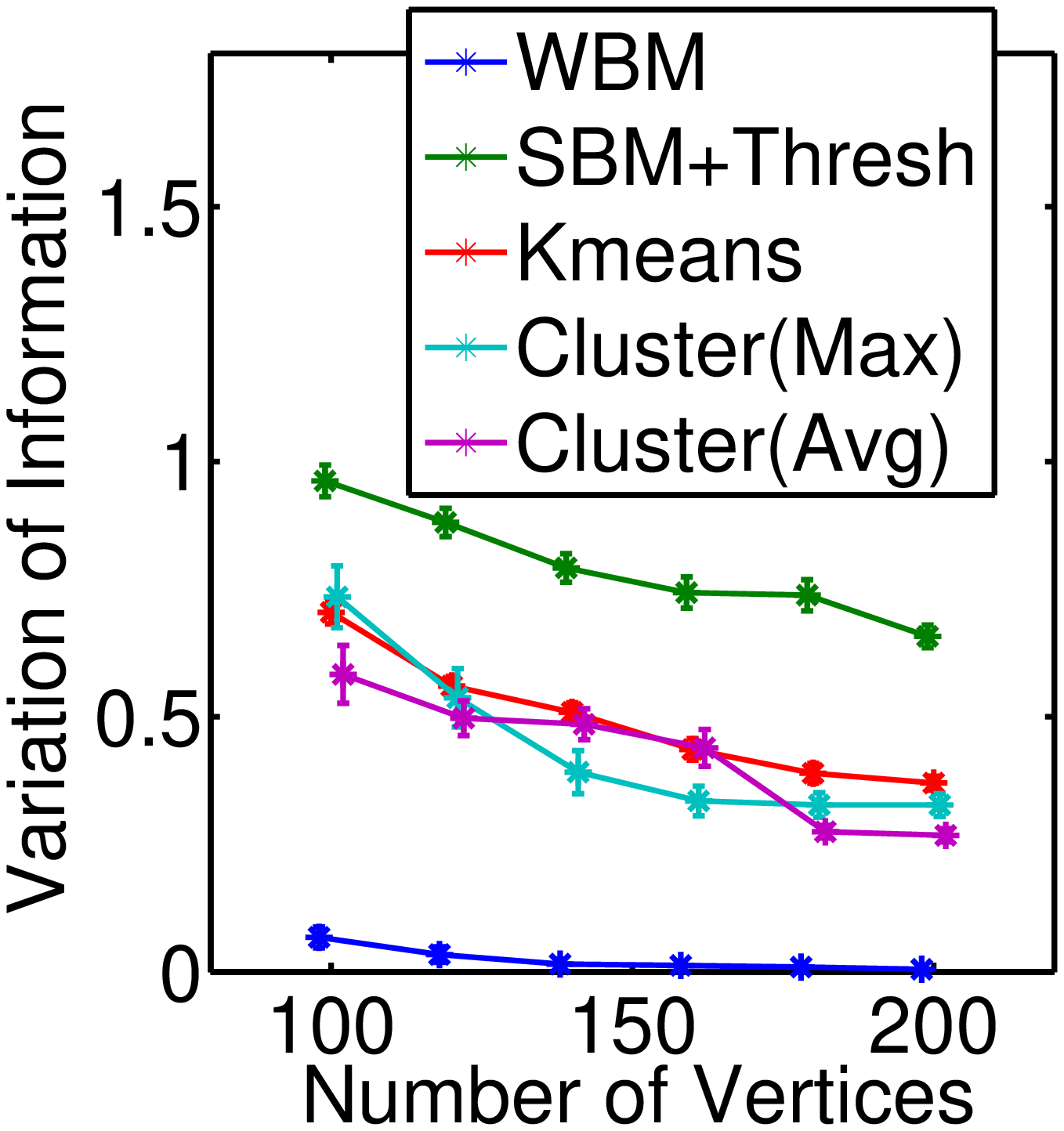}
			\label{fig:4}
			}
		\caption{Results of fitting the WSBM (blue) and other methods to our dense synthetic data. 
		(a) An example of a dense synthetic network with $n = 160$. 
		(b) Comparison of VI versus the parameter $k$, fixing the variance and $n = 160$. 
		(c) Comparison of VI versus the variance of the edges, fixing $k = 5$ and $n = 160$. 
		(d) Comparison of VI versus the size of the network $n$, fixing the variance and $k = 5$. 
		Points in (b,c,d) are averaged over 30 generated datasets. SBM with thresholding and K-means are averaged over 100 trials for each dataset for different thresholds.  
		}
		\label{Fig:Normal}
\end{figure*} 


We compare the WSBM against several alternative methods for recovering latent block structure. Our goal is to demonstrate that the classic SBM after applying a single threshold to all edge weights may miss important structure and that the WSBM can be used to explicitly evaluate the accuracy of inferring latent block via thresholding. We also include k-means clustering and hierarchical clustering to show that the weighted behavior the WSBM captures is different.
 
To demonstrate how the WSBM can find structure other methods may miss, we use synthetically generated dense graphs with $n$ vertices divided into $k^{*}= 5$ heterogenous blocks; the weights of each edge bundle are Normally distributed with bundle-specific parameters (see Fig.~\ref{fig:1}). This $5$-block model is a weighted variation of Newman's four-group test for unweighted graphs~\cite{newman_finding_2004}.
We then vary three model parameters---graph size $n$, variance of the edge weight distributions, and number of blocks we fit to the data---and measure the accuracy of the inferred block structure. Varying the graph size corresponds to consistency, varying the variance shows the performance in high-noise settings, and varying the number of blocks corresponds to robustness.



We characterize the accuracy of the recovered block structures using the variation of information (VI) ~\cite{meila_comparing_2007}, a standard metric for such tasks. The VI is a mathematically principled, information theoretic metric for the distance between the inferred and true assignment (vertex labels). Let $P$ denote the true block structure and $Q$ be our estimate. Then $\textrm{VI}(P,Q) =  H(P\,|\,Q) + H(Q\,|\,P)$, with $H(P\,|\,Q)$ being the conditional entropy. When $Q=P$ and we recover the true structure exactly, $\textrm{VI}(P,Q)=0$. One nice property of VI is that it increases only modestly when $Q$ differs from $P$ mainly by splitting or dividing blocks.



Under all test settings, the WSBM outperforms the alternatives (Fig.~\ref{Fig:Normal}b--d). As edge-weight variance increases, all methods have decreased performance, but the WSBM fails most gracefully.
As the graph size $n$ increases, all methods perform better, with the WSBM performing best by far. And, when varying the number of blocks we infer, all methods perform better when $k\approx k^{*}$, but only the WSBM correctly recovers the latent structure at $k = 5 = k^*$, which is the value selected under model selection using Bayes factors. Additionally, the WSBM fails gracefully when $k > k^*$.

\begin{figure}[h]         
	\centering
		\subfigure[Small Variance]{
			{\includegraphics[width = .21\textwidth]{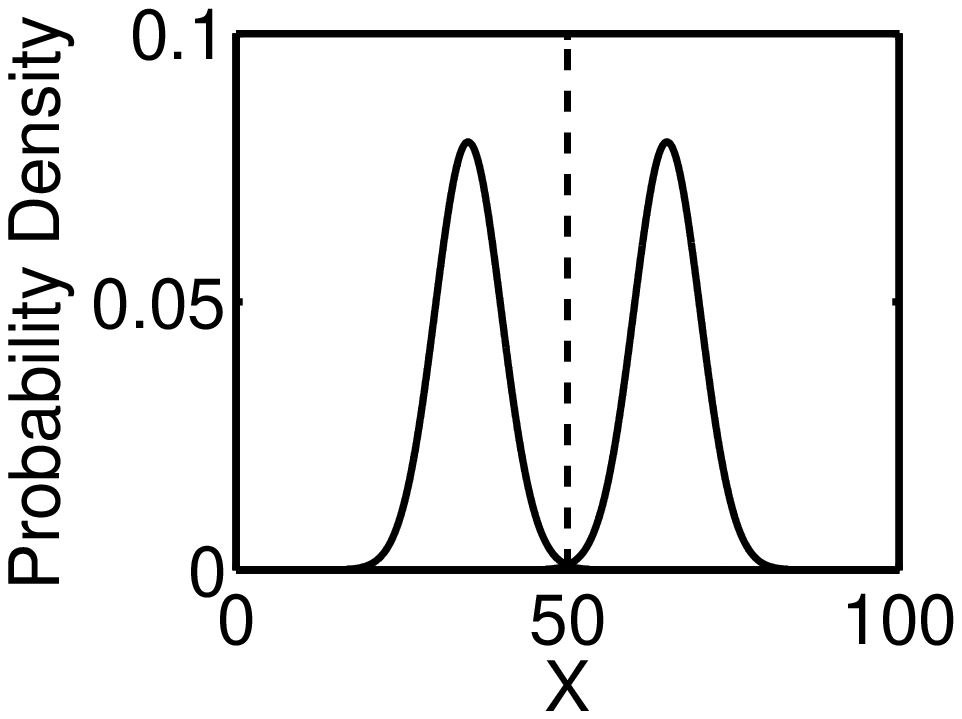}}
			\label{fig:v1}
			}
		\subfigure[Large Variance]{
			\includegraphics[width = .21\textwidth]{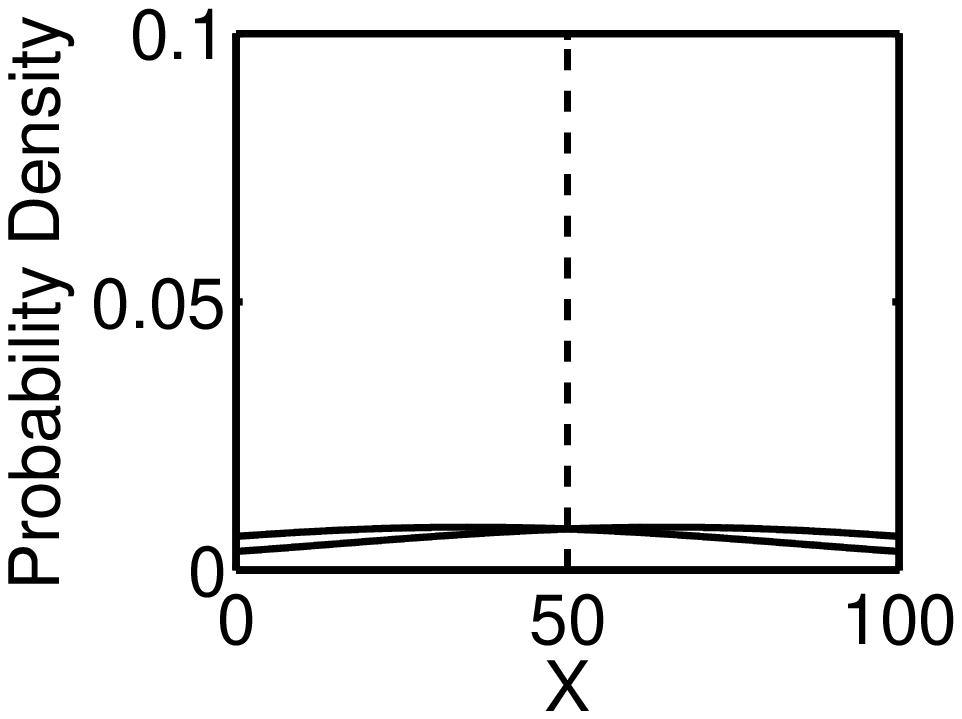}
			\label{fig:v2}
			}
		\caption{ The probability density plots (pdfs) of two pairs of normal distributions. In both figures, the distributions are centered at $x = 35$ and $x = 65$, but differ in variance and post-threshold probability. (a) The variance is $25$ and the probabilities of exceeding the threshold $x = 50$ are $0.001$ and $0.999$ respectively. (b) The variance is $2500$ and the probabilities of exceeding the threshold $x = 50$ are $0.380$ and $0.620$ respectively. }
		\label{fig:v}		
\end{figure}

Thresholding with the SBM performs poorly in all tests, because choosing a universal weight threshold destroys information about the latent block structure. 
Thresholding converts the original weights into a Bernoulli distribution with parameter equal to the probability of exceeding the threshold.
This effect is substantial whenever distinct blocks exhibit similar weight distributions.
If the two blocks' distributions are similar (Fig.~\ref{fig:v}), the SBM with thresholding typically finds only one block because the probabilities of exceeding the threshold are too similar.
In this case, thresholding confuses latent differences with Bernoulli sampling noise, and the SBM merges blocks that are distinct.
With well-separated weight distributions and an optimal threshold, the SBM may find correct structure. However, selecting the `optimal' threshold is a challenging problem itself.
Because a threshold will impact different edge bundles differently, a single `optimal' threshold may not, in fact, exist. 

As a result, when 
$k > k^*$, the SBM with thresholding tends to under-fit the data, leading to very poor results.
In contrast, the WSBM, having no thresholds, utilizes the complete weight information and performs well even when given more flexibility than the underlying data require.

The performance of k-means and hierarchical clustering is particularly poor for increasing edge-weight variance, when the signal-to-noise ratio is low. These methods over fit the data less than the classic SBM when given $k>k^{*}$, but they still perform more poorly than the WSBM. The reason for this difference is our particular choice example. The k-means algorithm uses principle component analysis, which suffers in high variance settings. Similarly, hierarchical clustering focuses on only intra-block behavior (the blocks on the diagonal) and misses out on inter-block behavior.   

\section{Discussion} 
The weighted stochastic block model we introduce here generalizes the classic stochastic block model to the important case of edges with weights drawn from an exponential family distribution. This generalization presented several technical challenges, which we solved using a Bayesian approach to develop a variational Bayes algorithm for dense graphs. 
This model accurately recovers latent block structure under a wide variety of conditions, and performs substantially better than simple alternatives. These results demonstrate that applying a threshold to edges weights before applying the unweighted SBM is generally unreliable.

The WSBM can be naturally generalized in several potentially useful ways. 
For sparse graphs, we have developed a scalable belief-propagation algorithm, to be presented in future work. 
It could also be extended to mixed membership~\cite{airoldi_mixed_2008} or, in the sparse case, to allow degree heterogeneity~\cite{karrer_stochastic_2011}. Stochastic variational inference has shown promising results for scaling in the mixed-membership SBM, and this technique could also be adapted to the WSBM~\cite{gopalan_scalable_2012}. Finally, an interesting question is the extent to which utilizing weight information modifies the phase transition in the detectability of latent block structure, which is known to exist in the classic SBM~\cite{decelle_phasetransition_2011}.

\subsection*{Acknowledgements}
We thank D.\ Larremore for helpful conversations. We acknowledge financial support from Grant \#FA9550-12-1-0432 from the U.S.\ Air Force Office of Scientific Research (AFOSR) and the Defense Advanced Research Projects Agency (DARPA).



\bibliography{WSBM_05}
\bibliographystyle{icml2013}

\end{document}